\documentclass{article}
\usepackage{ijcai23}

\usepackage{times}
\usepackage{soul}
\usepackage{url}
\usepackage[hidelinks]{hyperref}
\usepackage[utf8]{inputenc}
\usepackage[small]{caption}
\usepackage{graphicx}
\usepackage{amsmath}
\usepackage{amsthm}
\usepackage{booktabs}
\usepackage{algorithm}
\usepackage{algorithmic}
\usepackage[switch]{lineno}
\usepackage{comment}

\usepackage{bbding} 
\usepackage{bm}
\usepackage{color}
\usepackage{booktabs}
\usepackage{multirow}
\usepackage[normalem]{ulem}
\useunder{\uline}{\ul}{}

\usepackage{bbding} 
\usepackage{bm}

\usepackage{color}
\usepackage{booktabs}
\usepackage[normalem]{ulem}
\useunder{\uline}{\ul}{}

\usepackage{hyperref}

\newcommand\blfootnote[1]{%
	\begingroup
	\renewcommand\thefootnote{}\footnote{#1}%
	\addtocounter{footnote}{-1}%
	\endgroup
}

\title{Mimicking the Thinking Process for Emotion Recognition in Conversation \\with Prompts and Paraphrasing}

\author{
Ting Zhang$^1$\and
Zhuang Chen$^2$\and
Ming Zhong$^1$\textsuperscript{*}\And
Tieyun Qian$^{1,3}$\textsuperscript{*}
\affiliations
$^1$School of Computer Science, Wuhan University\\
$^2$The CoAI group, DCST, Tsinghua University\\
$^3$Intellectual Computing Laboratory for Cultural Heritage, Wuhan University
\emails
tingzhang\_17@whu.edu.cn, 
zhchen-nlp@mail.tsinghua.edu.cn, 
\{clock, qty\}@whu.edu.cn
}

\begin{document}

\maketitle
\begin{abstract}
\blfootnote{* Corresponding authors.}Emotion recognition in conversation,  which aims to predict the emotion for all utterances,  has attracted considerable research attention in recent years. It is a challenging task since the  recognition of the emotion in one  utterance  involves many complex factors, such as the conversational context, the speaker's  background, and the subtle difference between emotion labels. In this paper, we propose a novel framework which mimics the thinking process when modeling these factors. Specifically, we first \emph{comprehend the conversational context} with a history-oriented prompt to selectively gather  information from predecessors of the target utterance. We then  \emph{model the speaker's background} with an experience-oriented prompt  to retrieve the similar utterances from all conversations.  We finally  \emph{differentiate the subtle label semantics} with a paraphrasing mechanism  to elicit the intrinsic label related knowledge.
We conducted extensive experiments on three benchmarks. The empirical results demonstrate the superiority of our proposed framework over the state-of-the-art baselines.
\end{abstract}

\section{Introduction}
Emotion Recognition in Conversation (ERC) is a crucial task for recognizing mental health problems. As reported by the WHO, 1 in every 8 people in the world lives with a mental health problem such as emotional regulation~\footnote{https://www.who.int/news-room/fact-sheets/detail/mental-disorders.}. With the goal of recognizing human emotions, ERC can help discover the negative emotions of speakers and identify potential individuals who may be experiencing mental health issues.
The ERC task is also a fundamental step towards human-like artificial intelligence (AI) \cite{PoriaMMH19}, and has played an important role in many areas that are beneficial to humans such as legal trials \cite{PoriaMMH19}, empathetic dialog systems \cite{MajumderHPLGGMP20}, health care systems \cite{PujolMM19}, and intelligent assistants \cite{KonigFMH16}.

Different from conventional emotion recognition tasks, the emotion of a target utterance in ERC is not self-contained, which indicates that we cannot predict the emotion merely by understanding the utterance itself. Instead, some supplementary information, such as the conversational context, and the speaker's  background, is required to accurately identify the emotion conveyed by the utterance. Moreover, the difference between emotion labels like `\emph{sadness}' and `\emph{frustrated}' is often subtle and needs to be carefully distinguished.

Current research direction is mainly towards the modeling of the conversational context without taking the speaker's  background into account. Various sequence-based models \cite{HazarikaPMCZ18,MajumderPHMGC19,dialogueCRN} and graph-based models \cite{GhosalMPCG19,DialogXL,SGED,LiYQ22,ShenWYQ20} built upon pre-trained language models (PLMs) are developed to model contextual interactions between utterances.
There is also a growing trend in employing external commonsense knowledge \cite{ATOMIC} to enrich utterance representations~\cite{GhosalMGMP20,TODKAT} or facilitate emotion transition over the conversation graphs~\cite{SKAIG,CauAIN}.
Simply fusing features through network structure falls short of exploiting the knowledge capacity~\cite{liu2021pre,BrownMRSKDNSSAA20,GTPUnderstand} of the PLMs, and thus a more recent method CISPER~\cite{CISPER} leverages the prompt-learning paradigm for this purpose.
However, CISPER uses the same prompt for all utterances in a dialogue without considering their relations to the target utterance and its speaker.

In this paper, we propose a novel conversational emotion recognition framework which mimics the thinking process of a human being.
To understand the emotion conveyed by the target utterance, human beings typically go through the following questions step by step.
\begin{itemize}
\item[1)] \textit{What does the speaker say?}
People need first to locate and then read the target utterance to understand the utterance itself.
\item[2)] \textit{What is the influence of the conversational context on the speaker?}
The conversational context may exert a strong influence on  the speaker, so it is necessary to obtain relevant information from the dialogue history.
\item[3)] \textit{What is the speaker's background?}
People need to learn the speaker's background since the speaker often draws experience in similar situations to express an attitude to a particular utterance.
\item[4)] \textit{How does the speaker feel?}
People need to differentiate the semantics of the emotion labels for a precise emotion understanding.
\end{itemize}

To realize the above thinking process, we present a \textbf{m}ulti-\textbf{p}rompt and \textbf{l}abel \textbf{p}araphrasing (MPLP) model for emotion recognition in conversation.
Our model consists of two stages. At the first stage, the model is trained to understand 1) \textit{What does the speaker say?}. An utterance along with its surrounding context are fed into a PLM for the utterance encoding. 
In the second stage, the model is further trained to better identify the emotions with a thorough comprehension of 2) 3) 4).
While 2) is the focus of  all previous methods yet the  existing prompt based approach does not handle it well, 3) and 4) are largely neglected by current studies. Our effort is devoted to addressing these issues.

To be specific, to perceive 2) \textit{What is the influence of the conversational context on the speaker?}, we encode the speaker-related information and the history-influenced emotion into \textbf{a history-oriented prompt} to comprehend the conversational context.
To model 3) \textit{What is the speaker's background?}, we retrieve similar utterances seen in the training set and convert these utterances into \textbf{an experience-oriented prompt} to capture the speaker's task-specific experience.
To have a deep understanding of 4) \textit{How does the speaker feel?}, we design an auxiliary generation task with the help of  \textbf{label paraphrasing}  from SentiWordNet~\cite{BaccianellaES10} to distinguish the subtle semantics of different emotion labels.

In summary, the contributions of this work are threefold. Firstly, we point out the problem of inadequate coverage of the human thinking process in existing methods. Secondly, we propose a multi-prompt and label paraphrasing model to mimic this process in a comprehensive way.
Lastly, We demonstrate the effectiveness and the working mechanism of our proposed model via extensive experiments on three commonly-used datasets~\footnote{Our code and data are available at \url{https://github.com/NLPWM-WHU/MPLP}.}.

\begin{figure*}
    \centering	
    \includegraphics[width=0.9\linewidth]{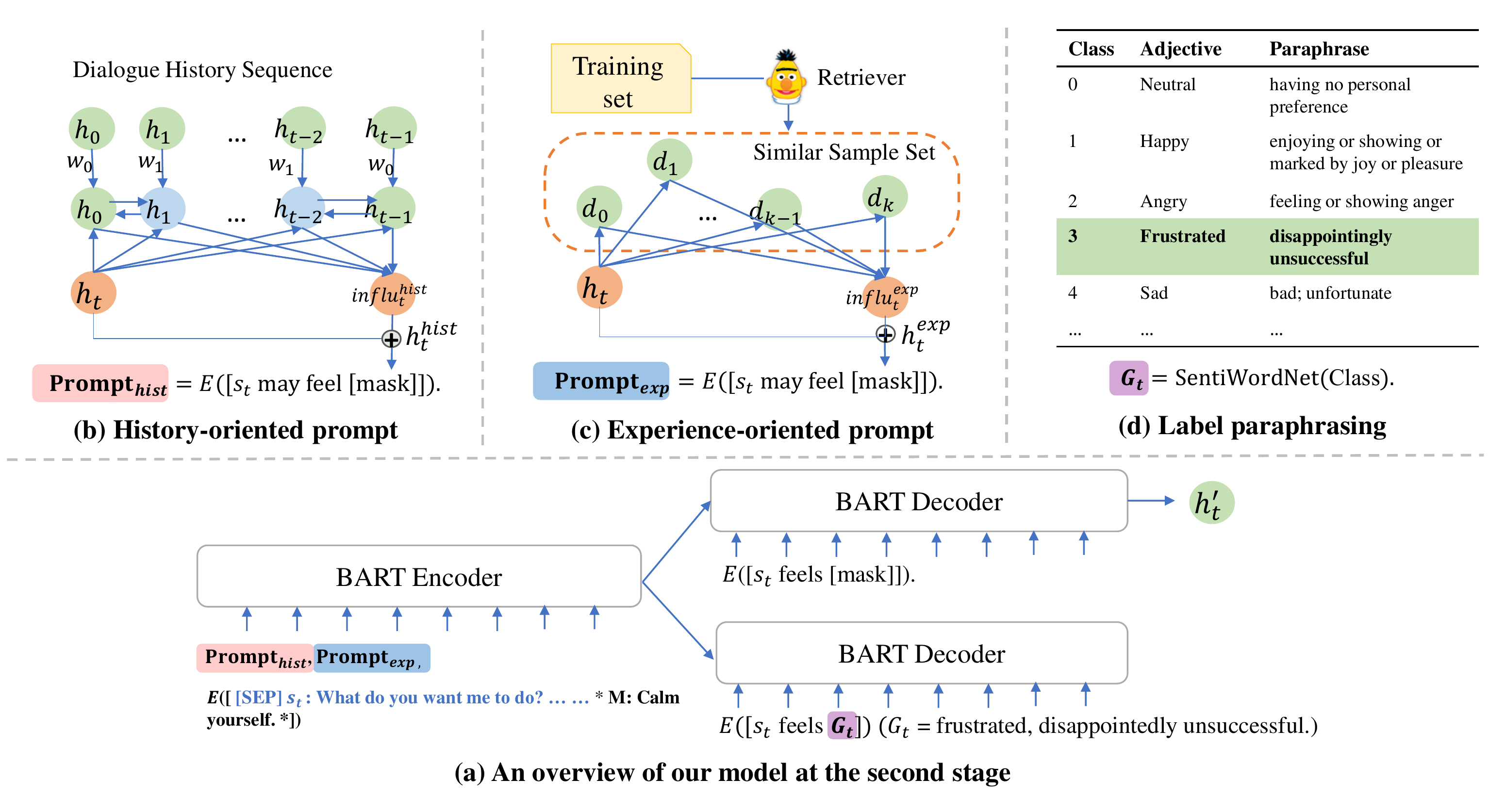}
	\caption{An overview of our model. The first stage for utterance understanding is conventional and thus omitted, and we only present the structure used at the second stage (a). The utterance representations from the first stage are used to construct the history-oriented prompt (b) and the experience-oriented prompt (c). The label paraphrases from the SentiWordNet are used for the auxiliary generation task (d).}
	\label{fig:fig_model}
\end{figure*}

\section{Related Work}
\subsection{Emotion Recognition in Conversation}
Most existing approaches design sequence-based or graph-based models to tackle the problem of context modeling. ICON \cite{HazarikaPMCZ18} uses GRUs to model the self- and inter-speaker emotional effects. DialogueRNN \cite{MajumderPHMGC19} keeps track of the individual party states by several GRU models.
DialogueCRN \cite{dialogueCRN} uses LSTM modules to retrieve and integrate contextual emotional clues iteratively.
DialogueGCN \cite{GhosalMPCG19} models speakers' dependency by applying graph neural networks to a neighbor graph. DAG-ERC \cite{ShenWYQ20} and SGED \cite{SGED} treat the conversation as an acyclic directed graph.
External knowledge is also widely used in ERC tasks. COSMIC~\cite{GhosalMGMP20} introduces commonsense knowledge during the sequence modeling procedure. TODKAT \cite{TODKAT} combines topic information to reduce the noise of commonsense.
SKAIG \cite{SKAIG} and CauAIN \cite{CauAIN} classify commonsense elements into different types to enhance emotion transition between utterances.
CISPER \cite{CISPER} encodes the conversation context and commonsense into prompts.
In addition to the model designing, some work~\cite{HCL,SupProto22,LiYQ22} adopts contrastive or curriculum learning strategies to get better results.

Overall, existing methods suffer from the issue of inadequate coverage of the human thinking process, and we realize this by mimicking the complete process with prompts and paraphrasing.

\subsection{Prompt and Paraphrasing}
Prompt-based learning is an emerging paradigm in natural language processing. To bridge the gap between the pre-training and fine-tuning, prompt-based methods modify the inputs by appending additional token sequences, which are helpful to elicit knowledge from PLMs \cite{BrownMRSKDNSSAA20,LesterAC21}.
Early models like GPT-3 \cite{BrownMRSKDNSSAA20}  use handcrafted task instructions and demonstrations to construct hard prompts. Recently, there has been a growing trend to explore the potential of continuous prompts  \cite{ACLLiL20,LesterAC21}. Paraphrasing is another new learning paradigm to transfer knowledge of a PTM by paraphrasing the key elements and generating a target sentence for the input sentence~\cite{MuellerKRMMZR22,ZhangD0YBL21}.

We introduce a history-oriented prompt, an experience-oriented prompt, and a label paraphrasing into our model, which can better leverage the power of PLMs for the downstream ERC task.

\section{Methodology}
\subsection{Problem Definition}
In ERC, a conversation is defined as a list of utterances ${u_1, u_2, ..., u_N}$, where $N$ is the number of utterances.
Each utterance $u_i$ consists of $n_i$ tokens, namely $u_i = {w_{i1}, w_{i2}, ..., w_{in_{i}}}$.
A discrete value $y_i \in Y $ is used to denote the emotion label of $u_i$, where $Y$ is the set of emotion labels.
Each utterance $u_i$ is associated with a speaker $s(u_i)$.
The objective of this task is to output the emotion label $y_t$ for a given query/target utterance $u_t$ based on its historical context $u_1, u_2, ..., u_{t-1}$ and the corresponding speaker information.

\subsection{Overview}
In this section, we present our  multi-prompt and label paraphrasing (MPLP) model to mimic the thinking process. 
The overview of our model is shown in Fig.~\ref{fig:fig_model}.
Our model consists of two stages. The first stage is for utterance understanding, i.e., ``1) \emph{What does the speaker say}?''. 
To this end, a PLM  is fine-tuned to produce initial representations for utterances. The resulting model is saved as the base model (denoted as MPLP$_b$).
The second stage is for the modeling of next three questions including ``2) \emph{What is the influence of the conversational context}?'', ``3) \emph{What is the speaker’s background}?'', and ``4) \emph{How does the speaker feel}?''.
Accordingly, we construct a history-oriented prompt and an experience-oriented prompt based on the initial utterance representations from the first stage.
Meanwhile, we perform an auxiliary task of label paraphrasing to leverage label semantics and fully elicit the lexical knowledge from the PLM.

\subsection{Utterance Understanding}
We adopt the generative PLM BART~\cite{LewisLGGMLSZ20} for utterance understanding.
We package the most recent $m$ utterances and their corresponding speaker names along with the target sentence into a  token sequence $C_t$, and feed it to the BART encoder. To distinguish the target utterance ($s_t, u_t$) from its context, a special token $ * $ is added at the beginning and the end of the input  utterance in the encoder. An emotional prompt $P_{t}$ is sent to the decoder to get the representation of target utterance. \textcolor{black}{$E$ denotes the embedding layer}:
\begin{align}
C_{t} & = [s_{t-m}, u_{t-m}, s_{t-m+1}, ..., *, s_t, u_t, * \ ]\\
P_{t} & = [s_t \text{\ feels \ [mask]} \ ]\\
\mathbf{H}_{t} & = \text{BART-Decoder}(\textcolor{black}{E(P_{t})}, \text{BART-Encoder}(\textcolor{black}{E(C_{t})})
\end{align}
The representation of the [mask] token in $\mathbf{H}_{t}$, denoted as $\mathbf{h}_{t}$, 
which reflects the underlying emotion understood by the model, will be used for training the model at the first stage with a cross entropy loss. After the first stage, we can obtain a preliminary understanding of the target utterance.

\subsection{History-oriented Prompt Construction}
Many studies \cite{HazarikaPMCZ18,MajumderPHMGC19,ShenWYQ20} have demonstrated the importance of historical information for the ERC task. However, the currently available prompt-based method CISPER~\cite{CISPER} simply constructs a shared prompt for all utterances in a dialogue. This hinders the model's ability to understand contextual information that is relevant to the target utterance. 
To address this issue, we propose a speaker-focused and history-oriented prompt generation method.

We generate a representation $\mathbf{h}_{i}$ for each historical utterance $u_i$ using the fine-tuned BART obtained by the first stage. To concentrate on the utterances that are highly relevant to the target utterance, we calculate the representation similarity as the importance measure:
\begin{align}
a^{hist}_i & = \frac{exp(\mathbf{W}^{hist}_{h}[\mathbf{h}_{i}; \mathbf{h}_{t}])}{\sum_{i=0}^{t-1}exp(\mathbf{W}^{hist}_{h}[\mathbf{h}_{i}; \mathbf{h}_{t}])}
\end{align}

Following \cite{ShenWYQ20}, a relation-aware feature transformation is applied to each historical utterance:
\begin{align}
\mathbf{h}_{i} & = \mathbf{W}^{hist}_{i} \mathbf{h}_{i}
\end{align}
where $\mathbf{W}^{hist}_{i} \in \{\mathbf{W}_{0}, \mathbf{W}_{1}\}$ is determined by whether a historical utterance $u_i$ is of the same speaker with the target utterance $u_t$. This helps to distinguish  the emotional effect of the current speaker and those of other speakers.

To make the emotional representation more contextualized, a Bi-LSTM module is applied to the historical sequence:
\begin{align}
\Tilde{\mathbf{h}}_{i} & = \text{Bi-LSTM}(\Tilde{\mathbf{h}}_{i-1}, {\mathbf{h}}_{i})
\end{align}

Finally, we aggregate the historical emotional information 
to obtain the influence $\mathbf{influ}^{hist}_t$  of the conversational context on the current speaker. $\mathbf{influ}^{hist}_t$
is further added to the original utterance representation to capture the emotional impact of the dialogue history:
\begin{align}
\mathbf{influ}^{hist}_t=\sum_{i=0}^{t-1}a^{hist}_i\Tilde{\mathbf{h}}_{i}, \ \mathbf{h}^{hist}_t = \mathbf{influ}^{hist}_t + \mathbf{h_t}
\end{align}

Further, to allow the PLM to better utilize  the target utterance-related history information, we construct a history-oriented prompt  by replacing the embedding at the [mask] position
in the original \textcolor{black}{$E([s_t \text{\ may \ feel \ [mask]}])$} prompt with $\mathbf{h}^{hist}_{t}$, and denote the resulting prompt as $\text{Prompt}_{hist}$:
\begin{align}
\text{Prompt}_{hist} & = \textcolor{black}{E([s_t \text{\ may \ feel}]) \ \mathbf{h}^{hist}_{t}}
\end{align}

$\text{Prompt}_{hist}$ is appended to the input of the encoder at the second stage. During the training procedure, the history-oriented prompt is updated by dynamically selecting historical relevant information to continuously enhance the model.

\subsection{Experience-oriented Prompt Construction}
The speaker's background is also crucial to determine his/her attitude in the conversation.
In particular, in multi-party conversations where the conversational context is less coherent, a speaker depends more on his/her experience in similar situations to facilitate conversation. In this section, we propose an experience-oriented prompt to encode the speaker's task-specific background.

We consider the training set, which has been seen by the model at the first stage  as the speaker's task-specific background, and retrieve similar samples in it to build the experience-oriented prompt. To find similar samples, we use a text retriever such as BERTScore ~\cite{ZhangKWWA20} and BM25~\cite{RobertsonW94}
to calculate the similarity between the target utterance $u_t$ and an utterance $u_d$ in the training set. The utterances with the top-$k$ similarity are chosen as the similar sample set $D$.
\begin{align}
    sim(u_t, u_d) = \text{Retriever}(u_t, u_d)
\end{align}

By now, we have calculated the \textit{text similarity} for the utterances themselves. However, as pointed out by previous work~\cite{LiYQ22}, even emotions of the same expression can vary dramatically in different context. To model the \textit{similar situation} more precisely, we calculate the \textit{context-influenced text similarity} between the similar samples and the target utterance, which is implemented by element-wise product of the utterance representations followed by a linear transformation. 
Since at the first stage, the local context is partially incorporated into the utterance encoding procedure, we consider their similarity as an indicator of the \textit{context-influenced text similarity} between utterances:
\begin{align}
    a^{exp}_j & = \frac{exp(\mathbf{W}^{exp}_{h}[\mathbf{d}_{j} \odot \mathbf{h}_{t}])}{\sum_{j=0}^{k}exp(\mathbf{W}^{exp}_{h}[\mathbf{d}_{j} \odot \mathbf{h}_{t}])} \\
\mathbf{influ}^{exp}_t&=\sum_{j=0}^{k}a^{exp}_j\mathbf{d}_{j}, \ \mathbf{h}^{exp}_t=\mathbf{influ}^{exp}_t + \mathbf{h}_t
\end{align}

After the above two steps, we can use the training samples which are similar to the current utterance and have similar context as our  \textit{prior experience} to get a deep  understanding of the speaker's background.
Similar to the previous section, we use $\mathbf{h}^{exp}_t$ to construct the experience-oriented prompt:
\begin{align}
\text{Prompt}_{exp} & = \textcolor{black}{E([s_t \text{\ may \ feel}])} \ \mathbf{h}^{exp}_{t}
\end{align}
$\text{Prompt}_{exp}$ is also appended to the input of the encoder at the second stage to  provide the speaker's background knowledge.

\subsection{Label Paraphrasing}
\label{label_repr_learning_sec}
The rich semantics of labels are indispensable for distinguishing the subtle difference between labels. They are also beneficial to  capture the text-label correlation.
In view of this, we perform an auxiliary label paraphrasing task to assist the main emotion recognition task.
We use the label names and their paraphrases in SentiWordNet 3.0~\footnote{Note that our model does not need paraphrase for inference.} to conduct the label paraphrasing task. To be specific, for a given label, such as \textit{sadness}, we map it to the corresponding adjective and generates the sense gloss $G_{t}$, which is the gloss of the most frequent sense~\footnote{We filter the paraphrases with both $PosScore$ and $NegScore$ equal to 0. Compared to the class category id and the adjective, the gloss contains more emotion-related words that are valuable for a deep label understanding.}. Finally, the input of the encoder at the second stage is denoted as $C^{\prime}_{t}$:
\begin{align}
I^{\prime}_{t} & = \text{Prompt}_{hist}, \ \text{Prompt}_{exp}, \\ \nonumber
 & E([\text{[SEP]} \ s_{t-k}, u_{t-k}, ..., *, s_t, u_t, *]) \
\end{align}
There are two generative targets $P_{t}$ and $G_t$ on the decoder side. These two targets are fed into the decoder for two passes to get $\mathbf{H}^{\prime}_{t}$ and $\mathbf{H}^{\prime}_{gt}$, which are used for emotion classification and label paraphrase generation, respectively:
\begin{align}
\mathbf{H}^{\prime}_{t} & = \text{BART-Decoder}(\textcolor{black}{E(P_{t})}, \text{BART-Encoder}(\textcolor{black}{I^{\prime}_{t}}))  \\
\mathbf{H}^{\prime}_{gt} & = \text{BART-Decoder}(\textcolor{black}{E([s_t \ \text{feels} \ G_{t}])},  \text{BART-Encoder}(\textcolor{black}{I^{\prime}_{t}}))
\end{align}

\subsection{Training and Prediction}
For the model training at the second stage, we take the representation of [mask] in $\mathbf{H}^{\prime}_{t}$ as the final representation $\mathbf{h}^{\prime}_t$, and apply a feed-forward neural network  to get the predicted emotion logits $p_{t}$ and the predicted label $\hat{y_{t}}$:
\begin{align}
    \mathbf{z}_{t} &= \text{GeLU}(\mathbf{W}_{H}\mathbf{h^{\prime}}_t + \mathbf{b}_{H})  \\
     p_{t} &= \text{Softmax}(\mathbf{W}_{z}\mathbf{z}_{t} + \mathbf{b}_{z})\\
     \hat{y_{t}} &= \text{argmax}_{e \in Y}(p_{t})
\end{align}

To fine-tune the model, the cross-entropy loss is used as the objective function:
\begin{align}
    L_{CE}(\theta) & = - \sum_{j=1}^{M} \sum_{i=1}^{N_j} \text{log}(p_{j, i} [y_{j,i}])
\end{align}
where $M$ is the number of conversations in the training set, $N_j$ is the number of utterances in the $j$-th dialogue, 
and $\theta$ is the collection of trainable parameters in our model.

The auxiliary loss for generating label paraphrases is calculated and added via weighted sum:
\begin{align}
    L_{GEN}(\theta) &= - \sum_{j=1}^{M} \sum_{i=1}^{N_j} \sum_{r=1}^{|G_{j,i}|}\text{log} p(g_{r+1} | g_r, \theta) \\
    L(\theta) & = L_{CE}(\theta) + \alpha * L_{GEN}(\theta)
\end{align}
where $g_r$ denotes the $r$-th token in the label gloss $G_{j,i}$, and $\alpha$ is a balancing weight for the loss of the label paraphrasing task.

\section{Experimental Settings}
\subsection{ Datasets and Metrics}
We conduct experiments on three widely used ERC datasets, including MELD \cite{MELD}, IEMOCAP \cite{IEMOCAP}, and DailyDialog \cite{dailydialog}.

\textbf{MELD} is collected from the TV show \textit{Friends}. It consists of multi-party conversations and there are 7 emotion labels including \textit{neutral}, \textit{happiness},
\textit{surprise}, \textit{sadness}, \textit{anger}, \textit{disgust}, and \textit{fear}.

\textbf{IEMOCAP} is a multimodal dyadic conversation dataset where each conversation is performed by two actors. There are 6 types of emotion, namely, $neutral$, $happiness$, $sadness$, $anger$, $frustrated$, and $excited$.

\textbf{DailyDialog} is a large collection of daily dialogues. In each conversation, there are two speakers. Each utterance is classified as \textit{neutral}, \textit{happiness}, \textit{sadness}, \textit{anger}, \textit{surprise}, \textit{disgust}, and \textit{fear}. Over 83\% of utterances in DailyDialog are classified as \textit{neutral}.

Only the textual modal information is used in our experiments. We adopt the micro-averaged F1 excluding the majority neutral class for DailyDialog and the weighted-average F1 for other two datasets as metrics \cite{ShenWYQ20}.

\subsection{Baselines}
We adopt 10 state-of-the-art baselines and divide them into two types: with or without external commonsense knowledge during inference. Our model belongs to the latter type.

\textbf{DialogueRNN} \cite{MajumderPHMGC19} uses three GRUs to keep track of the speaker states, proceeding contexts, and proceeding emotion.

\textbf{DialogueGCN} \cite{GhosalMPCG19} utilizes a graph-based structure to model self- and inter-speaker dependency of the interlocutors within a conversation.

\textbf{DialogXL} \cite{DialogXL} modifies the  XLNet~\cite{XLNet}  with dialog-aware self-attention to capture useful intra- and inter-speaker dependencies.

\textbf{DAG-ERC} \cite{ShenWYQ20} exploits a directed acyclic graph to model the information flow from both long-distance and nearby context in a conversation.

\textbf{CoG-BART} \cite{LiYQ22} employs the BART-Large model, and augments it with supervised contrastive learning and response generation to facilitate dialogue understanding.

\textbf{COSMIC} \cite{GhosalMGMP20} is the first model that incorporates different elements of commonsense and leverages them to update conversation states.

\textbf{TODKAT} \cite{TODKAT} combines topic information to help the model choose commonsense that is more relevant to the topic of current conversation.

\textbf{SKAIG} \cite{SKAIG} builds a locally connected graph and classifies commonsense elements into present, past, and future types to enhance emotion transitions in the graph.

\textbf{CauAIN} \cite{CauAIN} treats commonsense as the cause of emotion  and adopts the attention mechanism to connect utterances.

\textbf{CISPER} \cite{CISPER} is the first to leverage prompt learning for ERC. The context and commonsense of the entire conversation are encoded into shared prompts for utterances in a dialogue.

\subsection{Implementation}
The training process is divided into two stages. At the first stage, we fine-tune the BART-Larget model for a batch size of 8 utterances. Following~\cite{LiYQ22}, the AdamW optimizer is adopted with a learning rate of 2e-5 with a linear scheduled warm-up strategy. Our model is trained 4, 10, and 4 epochs for MELD, IEMOCAP, and DailyDialog, and the maximum input text length is set to 128, 160, and 128, respectively. After that, we start the prompting and paraphrasing for an additional 1 epoch at the second stage. The size of retrieved similar samples and the paraphrasing loss ratio is set via grid search. We use BERTScore as the text retriever for MELD and IEMOCAP, and BM25 for DailyDialog. The results on the test set come from the best checkpoint in the validation set. All experiments are performed on a single GeForce RTX 3090 GPU and are averaged over 3 runs.

\begin{table}[t]
\small
\tabcolsep=0.08cm
\begin{tabular}{cllll}
\toprule
\multicolumn{1}{l}{\textbf{}}                                                                         & \textbf{Model}  & \textbf{MELD}     & \textbf{IEMOCAP}       & \textbf{DailyDialog}    \\ \midrule
\multicolumn{1}{c}{}                                                                                
& DialogueRNN & 63.61          & 64.76         & 57.32          \\
\multicolumn{1}{c}{}                                                                                 
& DialogueGCN & 63.02          & 64.91         & 57.52          \\
\multicolumn{1}{c}{}                                                                                 
& DialogXL & 62.41             & 65.94         & 54.93          \\
\multicolumn{1}{c}{}                                                                                 
& DAG-ERC*    & 63.37          & \textbf{67.10} & 58.25          \\
\multicolumn{1}{c}{\multirow{-4}{*}{\begin{tabular}[c]{@{}c@{}}without \\ commonsense\end{tabular}}} 
& CoG-BART                     & 64.81          & 66.18         & 56.29          \\
\midrule
\multicolumn{1}{c}{}                                                                                 
& COSMIC                             & 65.21          & 65.28         & 58.48          \\
\multicolumn{1}{c}{}                                                                                 
& TODKAT                             & {\ul 65.47 }   & 61.33         & 58.47          \\
\multicolumn{1}{c}{}                                                                                 
& SKAIG                              & 65.18          & {\ul 66.96}   & {\ul 59.75}    \\
\multicolumn{1}{c}{}                                                                                 
& CauAIN*                            & 65.15          & 64.29         & 57.08          \\
\multicolumn{1}{c}{\multirow{-5}{*}{\begin{tabular}[c]{@{}c@{}}with \\ commonsense\end{tabular}}}    
& CISPER*                            & 64.99          & 55.20         & 56.80           \\ \midrule
\multicolumn{1}{c}{Ours}
& MPLP$_{b}$                         & 65.46          & 64.85         & 57.06          \\
& MPLP                               & \textbf{66.51} & 66.65         & \textbf{59.92} \\ \bottomrule
\end{tabular}
\caption{The main comparison results. 
For the models that use the test set for the checkpoint selection, we re-implement their official code and use the validation set for checkpoint selection for fair comparison. The results are marked with * in this table. }
\label{table:main}
\end{table}

\section{Results and Analysis}
\subsection{Main Comparison Results}
The main comparison results are shown in Table~\ref{table:main}. In general, our MPLP model achieves the best performance on MELD and DailyDialog datasets, and its performance on IEMOCAP is also very competitive. 

\textbf{MELD}
The models based on commonsense knowledge perform better on this dataset. This suggests that commonsense knowledge can provide additional information and facilitate the understanding of the utterances. Moreover, our base model $\text{MPLP}_b$, which does not utilize commonsense knowledge, can  achieve almost the same result as the best baseline TODKAT, indicating the strong utterance understanding capability of the PLM itself. Finally, with our proposed prompts and label paraphrasing, our complete model  outperforms TODKAT by an absolute 1.04 F1 increase. This clearly demonstrates that our human-mimicking process can get  a more comprehensive understanding of dialogue utterances.

\textbf{IEMOCAP} On this dataset, the graph-based models such as DAG-ERC and SKAIG can produce good results. The main reason is that the conversations in IEMOCAP is extremely long \cite{ShenWYQ20}, requiring complex stacked graph structures to model the dependencies of distant dialogues.
Note that the performance of CISPER, which is also based on prompt,  is extremely poor on this dataset.  In contrast, our model closely follows behind DAG-ERC and SKAIG and  is better than all other methods, indicating that our proposed prompt and paraphrasing learning mechanism can compensate for the issue of complicated dialogue structure modeling.

\textbf{DailyDialog} It is hard to recognize emotion from daily dialogues, and thus all methods produce worse results on this dataset than those
on other two datasets. Nevertheless, our model outperforms the models without commonsense by at least an absolute 1.67 F1 increase. It also surpasses the models with commonsense and future  information, showing the enormous understanding capacity  of our proposed human-mimicking framework.

\begin{table}[t]
\small
    \begin{tabular}{@{}l|lll@{}}
    \toprule
    & \textbf{MELD}  & \textbf{IEMOCAP}  & \textbf{DailyDialog} \\ \midrule
    Full Model       & \textbf{66.51} & \textbf{66.65} & \textbf{59.92}       \\
    w/o Hist Prompt  & 65.76($\downarrow$0.75) & 65.28($\downarrow$1.37) & 58.60($\downarrow$1.32)  \\
    w/o Exp Prompt   & 65.77($\downarrow$0.74) & 65.39($\downarrow$1.26) & 59.38($\downarrow$0.54)  \\
    w/o Label Para.  & 66.00($\downarrow$0.51) & 65.81($\downarrow$0.84) & 59.52($\downarrow$0.40)  \\
    \bottomrule
    \end{tabular}
\caption{The results for ablation study.}
\label{table:ablation}
\end{table}

\begin{figure}[t]
    \begin{minipage}[t]{0.5\linewidth}
        \centering
        \includegraphics[width=\textwidth]{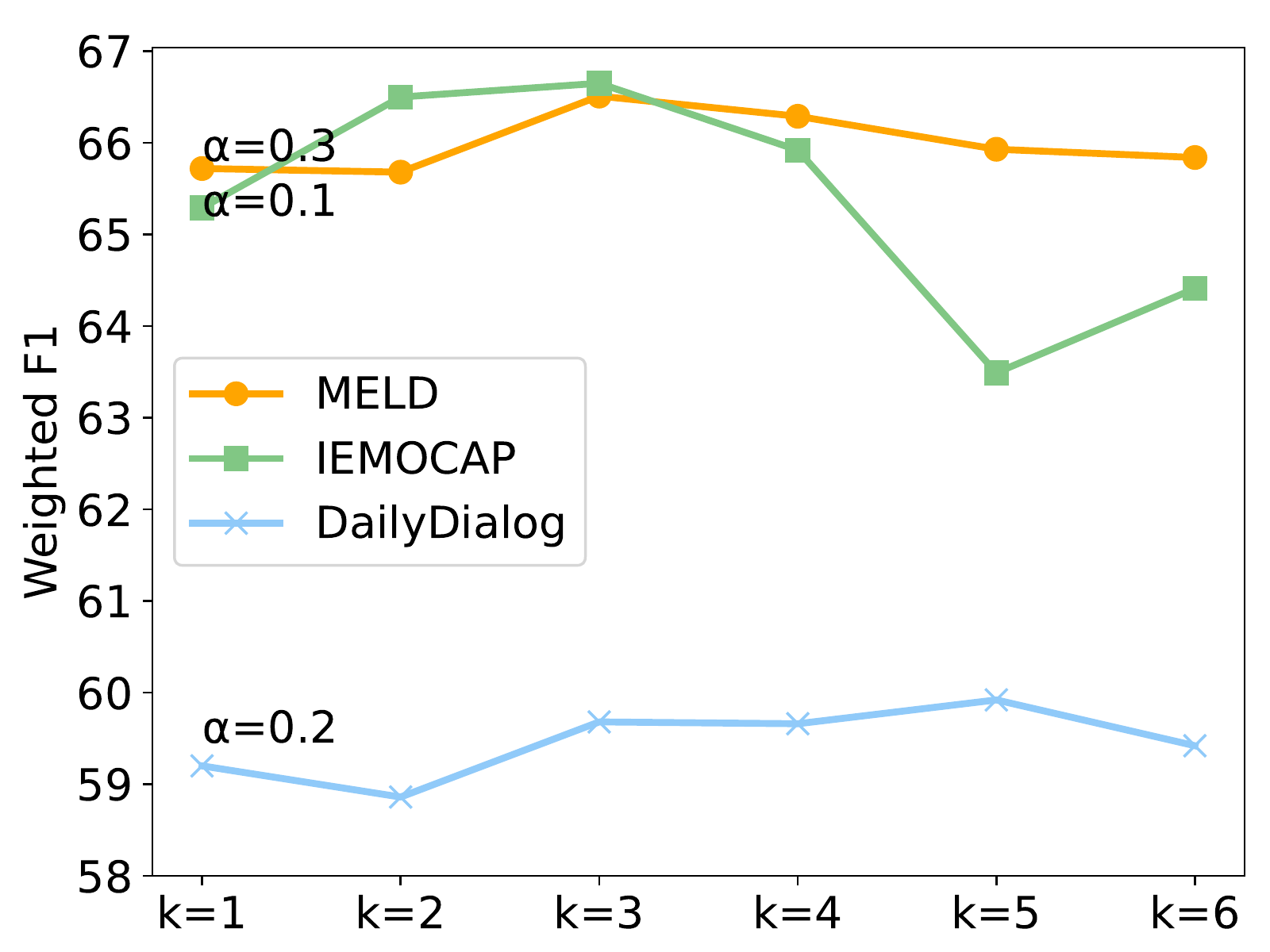}
        \centerline{(a)}
    \end{minipage}%
    \begin{minipage}[t]{0.5\linewidth}
        \centering
        \includegraphics[width=\textwidth]{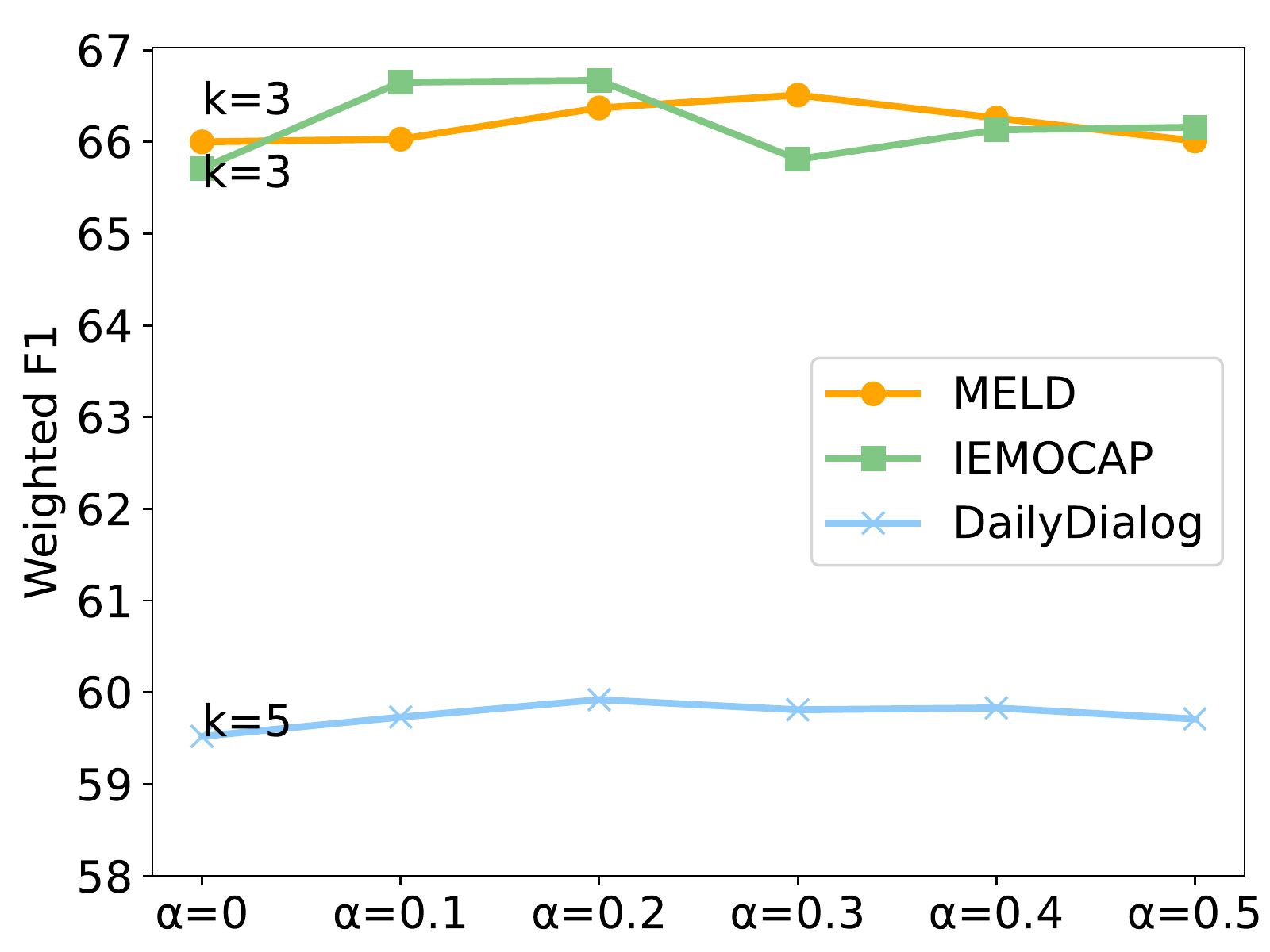}
        \centerline{(b)}
    \end{minipage}
    \caption{The results for parameter study.}
    \label{fig:parameter}
\end{figure}

\begin{table}[t]
\small
    \begin{tabular}{@{}l|lll@{}}
    \toprule
    \textbf{Model} & \textbf{MELD} & \textbf{IEMOCAP} & \textbf{DialyDialog}   \\ \midrule
    MPLP
    & { \textbf{66.51}} & { \textbf{66.65}} & { \textbf{59.92}}\\
    rep. Add
    & { 65.67($\downarrow$0.84)} & { 64.99($\downarrow$1.66)} & { 58.12($\downarrow$1.80)}\\
    rep. Concatenate
    & { 65.00($\downarrow$1.51)} & { 65.55($\downarrow$1.10)} & { 57.11($\downarrow$2.81)}\\
    \bottomrule
    \end{tabular}
\caption{The comparison results by using prompts and using features}
\label{table:prompt_compare}
\end{table}

\subsection{Ablation Analysis}
To verify the effectiveness of each component in our model, we conduct a series of ablation study, by removing the history-oriented prompt (denoted as \emph{w/o Hist Prompt}), the experience-oriented prompt (denoted as \emph{w/o Exp Prompt}), and the label paraphrasing task (denoted as \emph{w/o Label Para}) from the complete MPLP.
The results are shown in Table~~\ref{table:ablation}.
As we can observe, the performance drops on all datasets when each of the components is removed from the model. This proves the effectiveness of our proposed framework.

The history-oriented prompt yields the greatest impacts on three datasets. This is consistent with the findings in previous studies. Notably, the experience-oriented prompt makes almost the same contribution as the conversational
context on MELD and IEMOCAP. This demonstrates that the modeling of speaker's background, which is first attempt made by us for the ERC task,  is indeed essential  for the understanding of the target utterance in most cases.
The influence of experience-oriented prompt is not that significant on DailyDialog. The reason  might be that daily conversations are often diversified, and it is hard to trace the speaker's background from the training set.

Label paraphrasing shows the greatest impact on IEMOCAP. This might be due to that the emotion labels in this dataset are much ambiguous. 
For example, it is hard to distinguish $sadness$, $anger$, and $frustrated$  since these types of emotion are often mixed together, and thus label paraphrasing helps a lot in disambiguation.

\begin{table*}[t]
\small
\tabcolsep=0.015cm
\begin{tabular}{l|l|l|l|l}
\toprule[1pt]
\textbf{Conversational Context}  & \textbf{Similar Sample Set}  & \textbf{DAG-ERC} & \textbf{CISPER} & \textbf{Ours} 
\\ \hline
{......}                          & \begin{tabular}[c]{@{}l@{}} (1) Rachel: Mom, c'mon, stop worrying. (fear)\end{tabular}     &\textbf{}        & \textbf{}    & \textbf{}     \\ 
{Chandler:   I mean, that guy with the toe thing?  (anger)}  &  (2) Rachel: Hey, c'mon, cut it out. (joy)           &    &     & \\
{Chandler:   Who's he sleeping with? (neutral)}              &  (3) Rachel: Oh, c'mon. She's a   person,        
& sadness \XSolidBrush   & anger \XSolidBrush       & neutral \Checkmark      \\
{\ul \textbf{Chandler: Oh, c'mon Dora, don't be mad... (neutral)}}  &  you can do it! (joy)                                   
&                 &              &               \\
{......}     &       &      &       &       \\ \hline

{\ul \textbf{Monica: I   can't find garbage bags! (anger)}}     & \begin{tabular}[c]{@{}l@{}}(1) Chandler: I can't figure this  out! (anger)\end{tabular}     &   &    &    \\
{Rachel:   Oh, I think I saw some in here. (neutral)}       &  (2) Rachel: I can't watch! (fear)      & fear \XSolidBrush            & sadness \XSolidBrush        & anger \Checkmark         \\
{Monica:   What is it?! (surprise)}                         &  (3) Monica: I can't do it! (sadness)   &      &      &       \\
{......}         &        &       &       &   \\ \hline

{......}                                                   &  (1) Charlie: I was (surprise)     &       &       &      \\
{Phoebe:   Ohh, let me see it! Let me see your hand! (surprise)}     & (2) Chandler: I like her. (neutral)   &   &   &   \\
{Monica:   Why do you want to see my hand?    (neutral)}             & (3) Woman: I love your car. (joy)     & neutral \XSolidBrush         & neutral \XSolidBrush        & disgust \Checkmark      \\
{\begin{tabular}[c]{@{}l@{}}{\ul \textbf{Phoebe: I wanna see what's in your hand. I wanna see }} \\ {\ul\textbf{the trash. (disgust)}}\end{tabular}}      &       &          &      &               \\
{ …...}    &      &      &       &    \\
\bottomrule[1pt]
\end{tabular}
\caption{Case study on MELD. The target utterance in the conversational context is underlined.}
\label{table:case_study}
\end{table*}

\subsection{Parameter Study}
There are two parameters in our framework: the number of similar samples $k$  and the ratio of paraphrase generation loss $\alpha$. In this section, we investigate the impact of these two parameters. The results are drawn in Fig.~\ref{fig:parameter}.
We find that the trends on different datasets are similar with the change of parameters. They first rise to a peak and then fall gradually. A small $k$ does not provide enough experience. When $k$ is too large, the model is prone to introduce too much noise since more dissimilar sentences are added. The model with a small $\alpha$ can hardly learn label-semantics related information. If $\alpha$ is too large, the model may excessively emphasize the label semantics and ignores other factors.

\subsection{Using Prompts \emph{vs.} Using Features}\label{sec:pf}
We investigate the way in using the historical and background information, e.g., is prompt based learning a better way or can we simply use the same information as features?

To this end, we directly fuse the history  or experience influenced representations to the utterance embeddings instead of using them as prompts. To be specific, we add (denoted as \emph{rep. Add}) or concatenate (denoted as \emph{rep. Concatenate}) these representations to the decoder output for final classification. In the first case, the classification head of the first stage is reused. In the second case, we retrain a new classification head with more epochs for a fair comparison since the feature dimensionality increases. The results are shown in Table~\ref{table:prompt_compare}.

As can be seen, though the same representations from conversational context and the speaker's background are fused into the final embeddings, both the performance of \emph{rep. Add} and  that of \emph{rep. Concatenate} are consistently worse than our MPLP across all datasets. 
This indicates that the prompt based learning is a better way to inject relevant knowledge since the prompts are more understandable to a PLM than symbolic features.

\begin{table}[t]
\small
\tabcolsep=0.15cm
    \begin{tabular}{@{}l|lll@{}}
        \toprule
        \textbf{Model}  & \textbf{MELD} & \textbf{IEMOCAP} & \textbf{DailyDialog}   \\ \midrule
        MPLP  & { \textbf{66.51}} & { \textbf{66.65}} & { 59.92}\\
        rep. Special Token    & { 65.72($\downarrow$0.79)} & { 65.01($\downarrow$1.64)}  & { 59.75($\downarrow$0.17)} \\
        {rep. Label Adjective} & { 66.21($\downarrow$0.30)} & { 65.72($\downarrow$0.93)} & { \textbf{60.10}($\uparrow$0.18)} \\
        \bottomrule
    \end{tabular}
\caption{The impact of paraphrase design.}
\label{table:label_design}
\end{table}

\subsection{Paraphrase Design}
In this section, we examine the impact of paraphrase design on the model performance.
We employ three experimental setups. The default one is our proposed MPLP model in this paper, which uses the gloss of the target label in SentiWordNet as the generative target. The second one uses a special token for a label (denoted as \emph{rep. Special Token}). The third one adopts the corresponding adjective of the emotion label (denoted as \emph{rep. Label Adjective}) as the target. The results are shown in Table~\ref{table:label_design}.

We find that the performance of \emph{rep. Special Token}  variant  declines dramatically on all datasets. This is because the randomly initialized special tokens contain no semantics and cannot provide a proper guidance for the model.
Compared to the special token, the performance of \emph{rep. Label Adjective} variant is better, demonstrating the effectiveness of label understanding. The performance can be further improved by exploiting label glosses from SentiWordNet on MELD and IEMOCAP. However, the label's adjective works slightly better than the label paraphrasing on DailyDialog. This might be due to the diversity of the utterances in this dataset, where the
the most frequent paraphrase of the emotion label cannot well satisfy  such  requirement and the corresponding adjective provides a more precise meaning for the label.
Overall, label paraphrasing is helpful in most cases, yet its design  can be further optimized, which we leave for the future work.

\subsection{Case Study}
To have a close look at the impact of the proposed prompts and label paraphrasing mechanism in our framework,
We conduct a case study and show the results in Table~\ref{table:case_study}.

The first case shows the effectiveness of history-oriented prompt. 
DAG-ERC misclassifies the utterance as $sadness$, showing that it fails to capture the context. CISPER incorporates the context but it pays more attention on the utterance with $anger$ emotion.  
In contrast, our model can select  relevant information and make  the correct prediction of $neutral$. To further confirm the effect of the history-oriented prompt, we replace our prompt with feature concatenation (Sec.~\ref{sec:pf}), and find this variant also produces an $anger$ label.

The second case demonstrates the effectiveness of experience-oriented prompt.  As the first utterance in the dialogue, the target utterance does not have any history information. As a result, both DAG-ERC and CISPER are unable to accurately recognize its emotion. In contrast, our model can make a correct prediction based on its prior experience. As can be seen, there is a utterance  in the similar sample set which conveys the same emotion of $angry$ as that for the target utterance.

The third case emphasizes the importance of label paraphrasing, where no historical utterance or similar sample is related to the emotion of \textit{disgust}. DAG-ERC and CISPER directly classify the utterance as $neutral$. In contrast, the word ``trash'', which is associated with the label's semantics of ``disgust'', helps our model to make a correct prediction.

In summary, the history-oriented prompt, the experience-oriented prompt, and the label paraphrasing improve our model's capability to recognize emotion  in conversations.

\section{Conclusion}
We propose a novel  framework for the ERC task which mimics the thinking process of a human being.
We realize this process with a history-oriented prompt, an experience-oriented prompt, and the label paraphrasing mechanism, which
can improve the understanding of the conversational context, the speaker's background, and the label semantics, respectively.
We conduct experiments on three datasets, the results show that our method achieves competitive performance with the state-of-the-art baselines, 
proving the necessity of the modeling of human-thinking process, especially the understanding of the speaker's background, which has not been touched by existing studies.
The ablation study and in-depth analysis further confirm the importance and effectiveness of using prompts and paraphrasing in our framework.
\clearpage

\section*{Ethical Statement}
Emotion Recognition in Conversation has been a popular research topic in the academic community for a long time. The data and code used in our experiments are all open-source resources for research purposes.

\section*{Acknowledgements}
This work was supported by the grant from the National Natural Science Foundation of China (NSFC) project (No. 62276193). It was also supported by the Joint Laboratory on Credit Science and Technology of CSCI-Wuhan University.

\bibliographystyle{named}
\bibliography{ijcai23.bib}


\end{document}